\documentclass{article}

 \usepackage[preprint]{neurips_2026}

\usepackage[utf8]{inputenc} 
\usepackage[T1]{fontenc}    
\usepackage{hyperref}       
\usepackage{url}            
\usepackage{booktabs}       
\usepackage{amsfonts}       
\usepackage{nicefrac}       
\usepackage{microtype}      
\usepackage{xcolor}         
\usepackage{natbib}
\setcitestyle{numbers,square}
\usepackage{graphicx}
\usepackage{amsmath}
\usepackage{listings}
\usepackage{tabularx}

\title{\textsc{Intent2Tx}: Benchmarking LLMs for Translating Natural Language Intents into Ethereum Transactions}

\author{%
  Zhuoran Pan \\
  School of Computer Science\\
  Peking University\\
  Beijing, PA 100871 \\
  \texttt{zrpan25@stu.pku.edu.cn} \\
  \And
  Yue Li \\
  Taiyuan University of Technology \\
  Taiyuan, Shanxi, PA 030024 \\
  \texttt{liyue@tyut.edu.cn} \\
  \AND
  Zhi Guan \\
  Peking University \\
  Beijing, PA 100871 \\
  \texttt{guan@pku.edu.cn} \\
  \And
  Jianbin Hu \\
  Peking University \\
  Beijing, PA 100871 \\
  \texttt{hujianbin@pku.edu.cn} \\
  \And
  Zhong Chen \\
  Peking University \\
  Beijing, PA 100871 \\
  \texttt{zhongchen@pku.edu.cn} \\
}

\begin{document}

\maketitle


\begin{abstract}
The emergence of Large Language Models (LLMs) offers a transformative interface for Web3, yet existing benchmarks fail to capture the complexity of translating high-level user intents into functionally correct, state-dependent on-chain transactions. We present \textsc{Intent2Tx}, a high-fidelity benchmark featuring 29,921 single-step and 1,575 multi-step instances meticulously derived from 300 days of real-world Ethereum mainnet traces. Unlike prior works that rely on synthetic instructions, \textsc{Intent2Tx} grounds natural language intents in real-world protocol interactions across 11 categories, including diverse long-tail Decentralized Finance (DeFi) primitives. To enable rigorous evaluation, we propose an execution-aware framework that transcends surface-level text matching by employing differential state analysis on forked mainnet environments. Our extensive evaluation of 16 state-of-the-art LLMs reveals that while scaling and retrieval-augmentation enhance logical consistency and parameter precision, current models struggle with out-of-distribution generalization and multi-step planning. Crucially, our execution-based analysis demonstrates that syntactically valid outputs often fail to achieve intended state transitions, highlighting a significant gap in current "reasoning-to-execution" capabilities. \textsc{Intent2Tx} serves as a critical foundation for developing autonomous, reliable agents in intent-centric Web3 ecosystems. Code and data: \url{https://anonymous.4open.science/r/Intent2Tx_Bench-97FF/}
\end{abstract}


\section{Introduction}


In the Web2 ecosystem, the transition from explicit commands to goal-oriented interfaces is well-established; for instance, a user might say "book the fastest flight to London" rather than manually interacting with fragmented underlying APIs. LLMs are now catalyzing a similar shift in Web3, evolving into the primary interface that lowers the barrier to decentralized application (dApp) interaction. By moving from a manual transaction-based model to an "intent-centric" paradigm, LLMs allow users to express complex financial goals—such as "swapping ETH for USDC and then adding liquidity to a profitable pool"—in natural language.

However, translating natural language intents into executable on-chain transactions is significantly more complex than standard Web2 tool-calling. While existing AI research in function calling and API selection provides a foundation \cite{guo2024stabletoolbench, schick2023toolformer, patil2025berkeley}, Web3 introduces three unique hurdles: (1) Structural Heterogeneity, where diverse DeFi protocols use radically different router logics for similar purposes \cite{carter2021defi}; (2) Low Readability, as transaction signatures often involve nested tuples and raw hex data that lack the semantic clarity of RESTful APIs \cite{kitzler2023disentangling}; and (3) State-Dependency, where the validity of a transaction is inextricably linked to the real-time, irreversible state of the Ethereum Virtual Machine (EVM) \cite{hu2023evmtracer}.

Among benchmarks dealing with Web3 intents, two works are closely related to us, which include the Intent-Tx-18K dataset \cite{yaoauditable} and EVM-QuestBench \cite{yang2026evm}. However, the former is primarily designed to train intent-transaction consistency detectors; it relies on raw hex data from limited protocols (e.g., Uniswap, CoW \cite{xu2023sok}) without human-readable function signatures, rendering it unsuitable for generative tasks. The latter, while targeting transaction generation, provides inputs that resemble explicit, stepwise execution traces rather than goal-oriented intents. Such an approach fails to reflect real-world LLM applications where the objective is to abstract away, not dictate, the underlying technical steps. Consequently, there is a pressing need for a benchmark that bridges \textbf{high-level intent} with \textbf{readable and executable on-chain calls}.

To address these limitations, we introduce the \textbf{Intent-to-Transaction (\textsc{Intent2Tx}) benchmark}. We define the task as follows: given a high-level natural language intent (describing the what, not the how), the model must output a structured on-chain call comprising the contract address, function name, parameters, and value. Our benchmark encompasses two difficulty levels: \textbf{Single-step} (atomic operation that can be expressed in a single transaction) and \textbf{Multi-step} (a series of atomic operations requiring planning). The dataset is constructed from 300 days of Ethereum mainnet transaction logs (2025.03–2026.01), ensuring it reflects the most current and diverse protocol landscape. We have systematically evaluated the performance of state-of-the-art LLMs on this task, as well as the dataset’s utility as a high-quality fine-tuning corpus.

Our contributions are as follows:

\begin{itemize}
    \item First Intent-to-Transaction Benchmark: We present the first comprehensive benchmark that maps high-level intents to both readable and executable transactions, covering a wide array of DeFi protocols.

    \item General Extraction Pipeline: We propose a robust method for constructing intent-transaction pairs from real on-chain logs, providing a scalable way to synthesize high-quality training data from the blockchain's history.


    
    \item Execution-Aware Evaluation and Systematic Analysis: We introduce a simulation-based evaluation protocol using differential state analysis on forked mainnet environments, and conduct a comprehensive study of LLMs’ capabilities in intent-to-transaction generation, including the effects of retrieval, scaling, and cross-category generalization.
\end{itemize}

\section{Background and Related Work}

\paragraph{DeFi Primitives and the Intent-centric Paradigm.}
The DeFi ecosystem has evolved into a diverse landscape of financial primitives, spanning Decentralized Exchanges (DEXs), lending markets, liquid staking, and yield aggregators \cite{alamsyah2024review}. Traditionally, interacting with these protocols follows a transaction-based (imperative) model, where users must explicitly dictate the execution path—specifying contract addresses, function selectors, and ABI-encoded parameters. This requires a deep understanding of the EVM mechanics \cite{hildenbrandt2018kevm}. Recently, the intent-centric (declarative) paradigm \cite{goes2023anoma} has emerged to abstract this complexity. Unlike transactions that define how to execute (e.g., "call swapExactTokensForTokens on Uniswap V2"), intents focus on what the desired outcome is (e.g., "exchange 1 ETH for at least 2500 USDC"). While infrastructure like account abstraction (ERC-4337 \cite{buterin2021erc}) and off-chain solvers handle the execution of signed intents, a critical intelligence gap remains: the semantic translation of fuzzy natural language into the precise, structured calls required by these frameworks. Our work targets this gap, evaluating LLMs' ability to act as the primary translator across diverse Web3 operational categories.


\paragraph{Benchmark for Generating Structured Data.}
The task of transforming natural language into structured formats is a cornerstone of LLM evaluation. Current benchmarks predominantly focus on Text-to-SQL, code generation (e.g., Spider \cite{yu2018spider}, HumanEvalPro \cite{yu2024humaneval}), and tool-use/API calling (e.g., StableToolBench \cite{guo2024stabletoolbench}). In the context of blockchain transactions, however, the structured output space becomes substantially more demanding. Transaction schemas involve high-dimensional parameters, including 256-bit integers and deeply nested tuples \cite{lin2020modeling}, while also requiring strict adherence to domain-specific constraints such as slippage \cite{chemaya2024power} and deadline \cite{pacheco2023my}, where even minor deviations can invalidate execution. In addition, similar financial operations may map to heterogeneous function signatures across different DeFi protocols \cite{he2020characterizing}, further increasing the complexity of generation.
Our work extends structured data generation into the highly adversarial and high-stakes domain of Web3.


\paragraph{Conventions for Evaluating Executable Output.}
Reference-based metrics (e.g., BLEU) are increasingly replaced by execution-based evaluation in code synthesis to ensure functional correctness \cite{chen2021evaluating,austin2021program}. In blockchain settings, this paradigm is essential: transaction validity depends on the on-chain state at execution time (e.g., balances, protocol conditions, and temporal constraints), and even structurally correct outputs may fail or behave incorrectly outside their original environment. Most existing benchmarks, however, assume stateless, sandboxed functions and do not capture shared external state or irreversible actions \cite{guo2024api}. To address this, we introduce Differential State Analysis, which evaluates LLM outputs via simulated execution on a forked Ethereum mainnet and compares post-execution state changes (e.g., balance deltas) with ground truth.
\section{\textsc{Intent2Tx} Dataset}

\subsection{Overview}

\begin{figure}[h!]
    \centering
    \includegraphics[width=420bp]{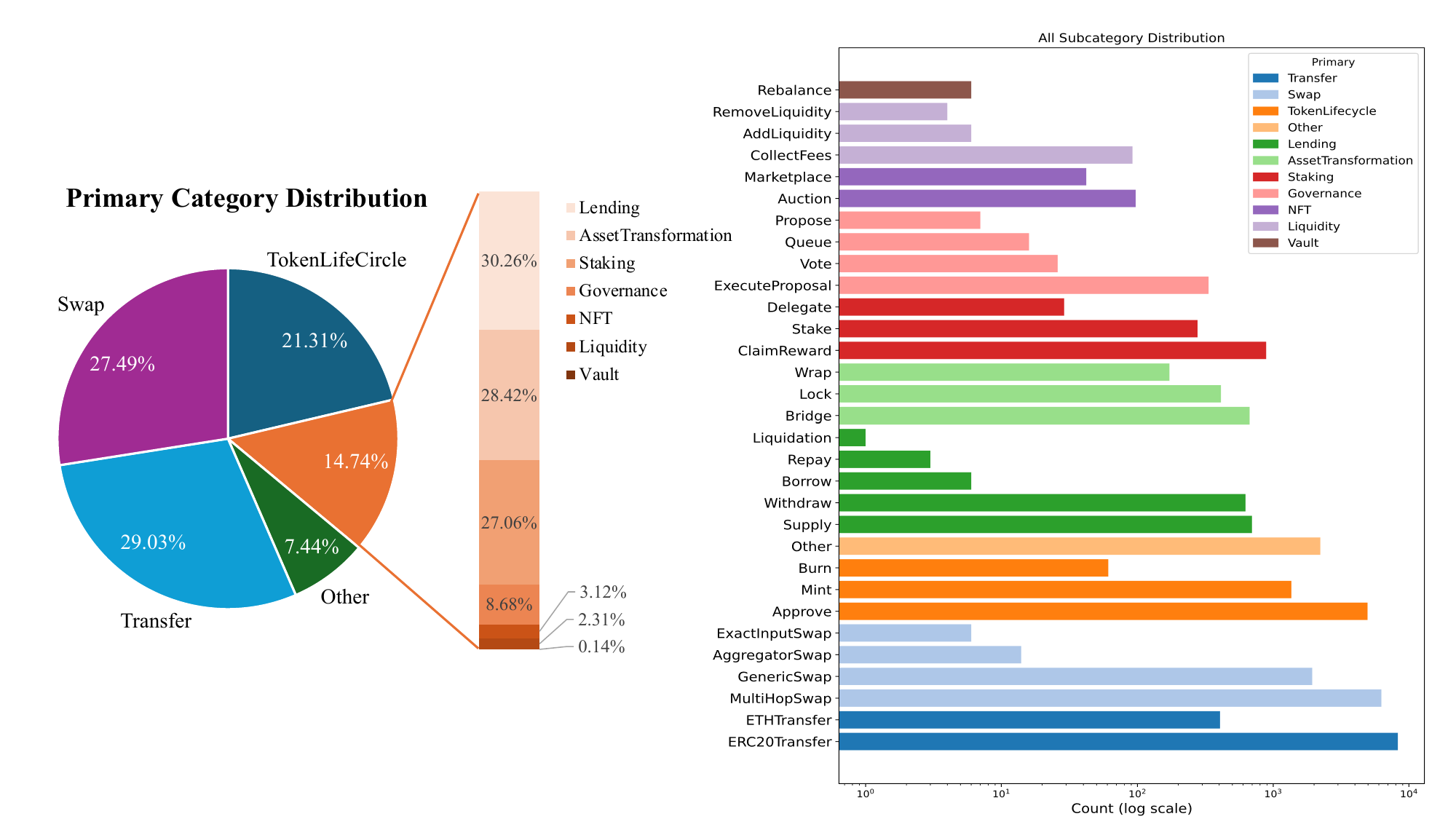}
    \caption{\textsc{Intent2Tx} dataset overview: size and (sub)category distribution. The distribution only shows the case of Single-step instances. Multi-step instanses are composed of combinations of Single-step as shown in Section \ref{data_cons_pipeline}, so no categories are defined.}
    \label{category_distribution}
\end{figure}

The \textsc{Intent2Tx} dataset is a large-scale benchmark designed to bridge the gap between natural language intents and executable blockchain transactions. It consists of \textbf{29,921} Single-step instances and \textbf{1,575} Multi-step instances, each meticulously derived from real-world Ethereum mainnet transaction logs spanning from March 2025 to January 2026.

Each instance in the dataset is structured as a mapping from a high-level user intent to an executable transaction schema. Taking single-step instances as example, the input is a natural language sentence describing a specific financial or operational goal (e.g., ``Swap 1,000 USDC for ETH with minimal slippage''). The corresponding output is a structured on-chain call which contains contract name, address, function, parameters and ETH value.
This structured format ensures that the generated output is not only human-readable but also directly simulation-ready on the EVM.

As illustrated in Figure~\ref{category_distribution}, \textsc{Intent2Tx} captures the multifaceted nature of the Web3 ecosystem. The primary category distribution is dominated by core DeFi activities: Transfer, Swap, and TokenLifeCircle. This aligns with the most frequent behaviors of retail users.

Beyond these dominant categories, the dataset exhibits a significant ``long-tail'' distribution, covering more complex protocols such as Lending, Staking, Governance, and Liquidity Management, which collectively represent the diverse utility of the Ethereum network. The subcategory distribution (Figure~\ref{category_distribution}, right) further reveals high-granularity actions like MultiHopSwap, Bridge, ClaimReward, and ExecuteProposal. By grounding our benchmark in 300 days of live on-chain history, \textsc{Intent2Tx} provides a high-fidelity simulation of real-world parameter distributions, varying protocol architectures, and evolving contract patterns, posing a rigorous challenge for LLM-based intent translation.

\subsection{Task Definition}

\textbf{Problem Formalization. }
The goal of the \textsc{Intent2Tx} task is to map a high-level natural language intent $I$ into a sequence of structured, executable blockchain transactions $t$. Formally, the task is defined as finding a mapping function $f(I) \to [ t_1, t_2, \dots, t_k ]$, where $k=1$ for atomic, single-step operations and $k>1$ for complex, multi-step workflows. Each transaction $t_i$ must contain all necessary metadata and parameters to be successfully executed on-chain.

\textbf{Transaction Schema. }
To ensure the output is both human-readable and machine-executable, we define a standardized JSON-based Transaction Schema for $t_i$:

\begin{footnotesize}
\begin{verbatim}
{
    "contract": string,             // contract name (e.g., "UniswapV2Router02")
    "contract_address": string,     // contract address (e.g., "0x7a250d5630...")
    "function": string,             // function name (e.g., "transfer")\n'           
    "params": {                     // map of parameters\n'
        "<param_name>": {
            "type": string,         // Solidity type (e.g., "address", "uint256")
            "val": any              // specific value inferred from the intent
        },
        ... // more parameters if needed
    },
    "value": float                  // Ether amount sent with the transaction (in ETH)
}
\end{verbatim}
\end{footnotesize}

This design prioritizes three key criteria: (1) Generality: It abstracts away the need for pre-loaded ABIs by explicitly including function names and types, making it protocol-agnostic; (2) Execution-Readiness: The inclusion of both \texttt{contract\_address} and typed \texttt{params} allows the output to be directly converted into raw calldata for simulation; (3) LLM-Friendliness: By using semantic names (\textit{contract}, \textit{function}) rather than raw hex codes, the schema leverages the pre-trained linguistic knowledge of LLMs to improve alignment.

\textbf{Task Difficulty. }
We categorize the benchmark based on whether an intent can be fulfilled within a single on-chain transaction:

Single-step: The intent can be expressed and executed using a single Ethereum transaction ($k=1$). This setting evaluates the model's ability to identify the correct contract function and infer precise parameters (e.g., an ERC20 transfer or a token swap).

Multi-step: The intent requires a sequence of multiple transactions ($k>1$), where intermediate steps must be correctly ordered and parameterized (e.g., \textit{Approve $\to$ Swap $\to$ Add Liquidity}). This setting tests the model's ability to decompose high-level goals into executable steps and handle inter-transaction dependencies.

\subsection{Data Construction Pipeline}
\label{data_cons_pipeline}

\begin{figure}[h!]
    \centering
    \includegraphics[width=400bp]{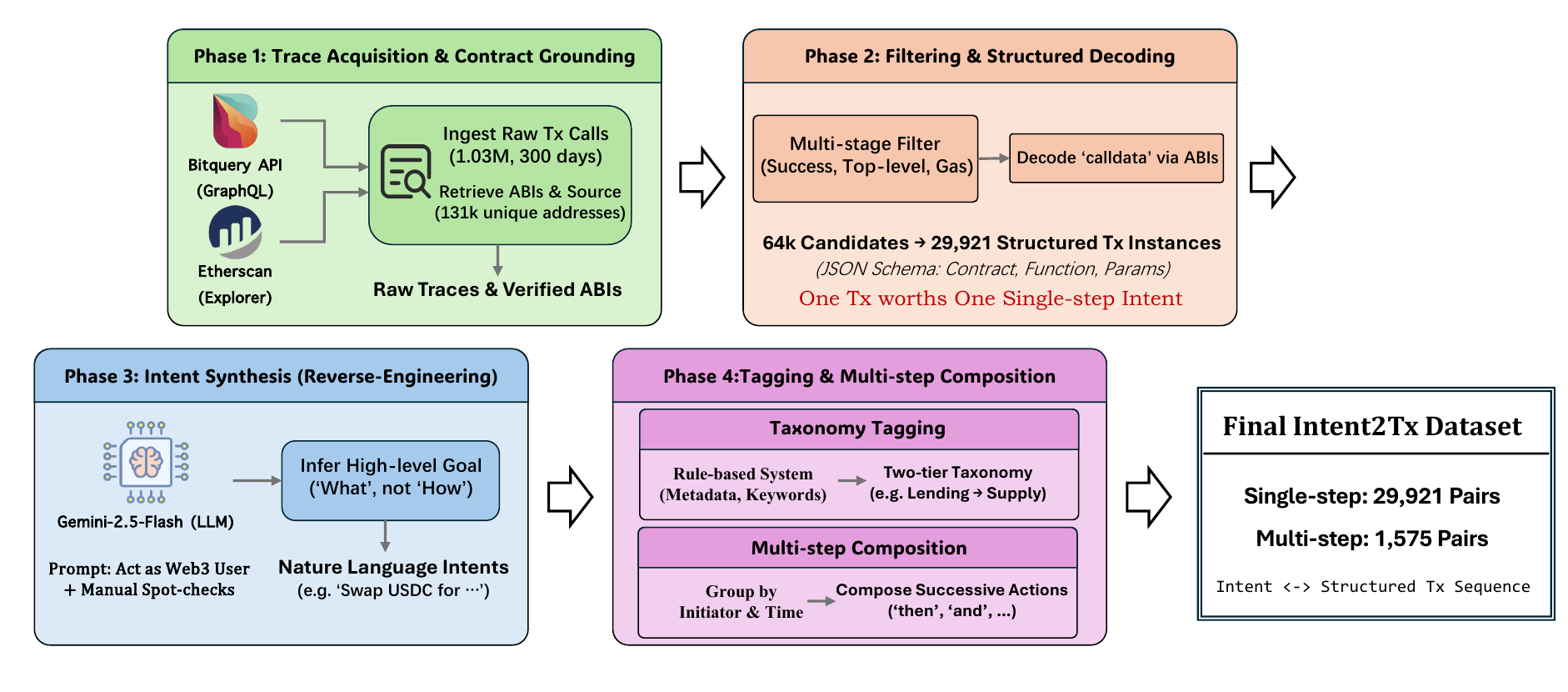}
    \caption{The four-phase data construction pipeline for the \textsc{Intent2Tx} benchmark, from raw on-chain traces to structured intent-transaction pairs.}
    \label{data_pipeline}
\end{figure}

The construction of \textsc{Intent2Tx} follows a rigorous four-phase pipeline, transforming raw, low-level blockchain traces into high-level, semantically rich intent-transaction pairs. This process ensures that the resulting benchmark is grounded in real-world economic activity while maintaining human-readability. The detailed data construction pipeline is shown in Figure \ref{data_pipeline}.

\textbf{Phase 1: Trace Acquisition and Contract Grounding. }
We initially ingested over 1.03 million raw Ethereum transaction calls spanning a 300-day window via the Bitquery GraphQL API. To ensure the transactions could be decoded into a structured format, we extracted 131,273 unique contract addresses and retrieved their corresponding source code and ABIs from Etherscan, a well-known blockchain explorer. This grounding step is crucial for moving beyond raw hex data and enabling the extraction of semantic function signatures.

\textbf{Phase 2: Filtering and Structured Decoding. }
Raw traces are often cluttered with system calls, failed transactions, and "spam" interactions with zero functional value \cite{zhang2020preventing}. We applied a multi-stage filter to retain only successful and top-level atomic calls that interacted with verified contracts and consumed enough gas. Utilizing the retrieved ABIs, we decoded the raw \texttt{calldata} of 64,424 candidate transactions into the \textbf{Transaction Schema} defined in the previous section. This resulted in 29,921 high-quality, structured transaction instances, mapping hex identifiers to human-readable function names and typed parameters.

\textbf{Phase 3: Intent Synthesis via Reverse-Engineering. }
To generate realistic user intents, we referred to the proven effective patterns of LLM participating in dataset construction \cite{patil2024gorilla, xu2023tool, liu2023makes}. We employed a "reverse-engineering" approach using \texttt{gemini-2.5-flash}. Rather than asking the model to simply describe the code, we prompted it to act as a Web3 user and infer the high-level goal (\textit{intent}) behind the structured schema. For common actions (e.g., Swaps), the model was instructed to use colloquial language; for niche DeFi protocols, it maintained technical precision. To ensure quality, we performed manual spot-checks on the generated intents to confirm they described the "what" (the goal) rather than the "how" (the specific function parameters). The prompt we obtained through multiple rounds of tuning and small-scale trials can be found in the Appendix \ref{prompt_1}.

\textbf{Phase 4: Taxonomy Tagging and Multi-step Composition. }
The final dataset was enriched with a two-tier taxonomy (e.g., \textit{Lending $\to$ Supply}). We implemented a rule-based tagging system based on contract metadata and function keywords to categorize the 29,921 \textbf{Single-step} instances. The labeling results are shown in Figure \ref{category_distribution}. To construct the \textbf{Multi-step} benchmark, we grouped transactions by the initiator's (\texttt{from}) address and execution timestamp. Successive actions performed by the same user within a short window (e.g., \textit{Approve} followed by \textit{AddLiquidity}) were composed into complex intents using natural conjunctions (``then'', ``and''), resulting in 1,575 sophisticated multi-step planning tasks. Examples of data from the benchmark are shown in Appendix \ref{data_example}.

\section{Experiments and Evaluation Framework}

To provide a comprehensive assessment of current AI capabilities in the Web3 domain, we evaluate a diverse suite of 16 LLMs. These models are selected based on their performance in general reasoning, code synthesis, and instruction following. Furthermore, we include specialized variants specifically optimized for programming tasks (e.g., GPT-5.2 Codex, Qwen3 Coder) to investigate whether code-specific pre-training translates to superior performance in transaction schema generation. The full list of evaluated models and their respective versions is detailed in Appendix \ref{llms_eval}.

\subsection{Research Questions and Experimental Settings}

\textbf{RQ1: How well do LLMs perform on \textsc{Intent2Tx}?}
We compare two inference settings: (1) \emph{Direct inference}, where the model generates transactions from the system prompt (Appendix \ref{prompt_2}), user intent, and an address book (Appendix \ref{address_book}) without in-context examples; and (2) \emph{Retrieval-augmented inference}, where the top-$k=3$ similar instances are retrieved as demonstrations. Similarity is computed via Jaccard overlap over <contract, function> pairs, with a fallback to down-weighted text similarity. All models are evaluated on both single-transaction and multi-transaction tasks under the same stratified sampling scheme.

\textbf{RQ2: How does training data scale affect performance?}
We conduct a scaling experiment to study how training data size affects generalization in intent-to-transaction generation. We fine-tune a Qwen3-14B model \cite{yang2025qwen3} using parameter-efficient LoRA \cite{li2023loftq} adaptation implemented in LLaMA-Factory \cite{zheng2024llamafactory} on progressively larger subsets of the training data, with sizes of 200, 800, 2k, 8k, 14k, 20k, and 28,776 instances.

To ensure a balanced distribution, all training subsets are constructed via stratified sampling over intent categories. The remaining 1,145 instances are held out as a fixed test set across all configurations.

Our hypothesis is that increasing the scale and diversity of training data improves the model’s ability to generalize to unseen intents and transaction compositions. This setup allows us to analyze scaling trends under a controlled data distribution and consistent evaluation protocol.

\textbf{RQ3: Can models generalize to unseen operation categories?}
We evaluate whether models fine-tuned on common operation categories can generalize to rarer and more complex categories. Specifically, we train on a subset of frequent categories, including \{Transfer, Swap, TokenLifecycle, Lending, Staking, AssetTransformation\}, and test on a disjoint set of less frequent categories: \{Governance, Liquidity, NFT, Vault, Other\}. The fine-tuning configuration follows the same setup as in Scaling experiment. By enforcing a strict category split, this experiment isolates cross-category transfer ability and evaluates whether models can compose previously learned primitives to handle unseen operation types.

\textbf{RQ4: How well do generated transactions execute on-chain?}
We note that multiple structured transaction sequences may achieve the same user intent, even if they differ from the reference output (e.g., invoking alternative contracts or equivalent functions). To account for this, we design a simulation-based evaluation framework that measures execution-level equivalence rather than exact match.

Given both the reference ($G$) and model-generated ($P$) transactions, we first extract the contract address, function, and parameters, and reconstruct the corresponding Ethereum transactions via ABI encoding. We then execute both transactions on a forked Ethereum mainnet using \texttt{anvil}\cite{tudorache2025detailed}, where the fork point is set to one block prior to the original transaction.

We evaluate two criteria: (1) \emph{executability}, i.e., whether the transaction can be successfully executed, and (2) \emph{state consistency}, defined as whether the resulting account-level state changes (e.g., token balance deltas) are consistent between $G$ and $P$, within a tolerance threshold ($<1\%$).

This framework enables a more faithful assessment of functional correctness by capturing semantic equivalence in execution outcomes, rather than relying solely on syntactic matching.

\subsection{Evaluation Metrics}

To assess the functional alignment between the prediction $P$ and the gold standard $G$, we define a multi-dimensional scoring framework. This framework evaluates the model’s performance across structural, logical, and parameter-level accuracy.

\textbf{Format Score ($S_{fmt}$)} measures basic instruction-following by verifying JSON validity: 
$$S_{fmt}(G, P) = \begin{cases} 1, & \text{if } P \text{ is a valid JSON object} \\ 0, & \text{otherwise} \end{cases}$$

\textbf{Logic Score ($S_{log}$)} evaluates the model's ability to identify the correct interaction target, including the contract address, function signature, and contract name:
$$S_{log}(G, P) = 0.4 \cdot \mathbb{I}(addr_P = addr_G) + 0.4 \cdot \mathbb{I}(func_P = func_G) + 0.2 \cdot \mathbb{I}(name_P = name_G)$$
where $\mathbb{I}(\cdot)$ is the indicator function. 

\textbf{Param Score ($S_{prm}$)} assesses the granularity of parameter extraction. Let $K_{all} = K_G \cup K_P \cup \{ Value_{eth} \}$ represent the union of standard and predicted parameter keys. The score is calculated as:
$$S_{prm}(G, P) = \frac{1}{|K_{all}|} \sum_{k \in K_{all}} s(k)$$
Individual parameter accuracy $s(k)$ is defined by a base score of 0.4 for key existence, with additional increments of 0.3 for correct type matching and 0.3 for value matching.

\textbf{Pass@1 ($S_{pass}$)} serves as a strict "Hard Match" indicator for end-to-end usability, requiring near-perfect logic and perfect parameter alignment:
$$S_{pass}(G, P) = \mathbb{I}(S_{log} \ge 0.8 \land S_{prm} = 1.0)$$

The \textbf{Final Weighted Score} for an instance $i$ combines these sub-metrics to reflect overall reliability:
$$S_{final} = \omega_1 S_{fmt} + \omega_2 S_{log} + \omega_3 S_{prm} + \omega_4 S_{pass}$$
Following empirical requirements for stable on-chain execution, we assign weights of $\omega_1 = 0.1, \omega_2 = 0.3, \omega_3 = 0.2$, and $\omega_4 = 0.4$.

For Multi-step tasks where $G$ contains $T$ sequential actions, the metrics are extended by taking the arithmetic mean of each step:
$$S_{metric\_avg} = \frac{1}{T} \sum_{t=1}^{T} S_{metric}(G_t, P_t)$$
The total performance is reported as the average score across all $N$ samples in the benchmark.
\section{Results and Analysis}

\subsection{Retrieval Augmentation Significantly Improves Performance (RQ1)}

\begin{table*}[h!]
\centering
\scriptsize
\setlength{\tabcolsep}{3pt}
\caption{Evaluation results across direct and retrieval settings. Each setting reports format (F: $S_{fmt}$), logic (L: $S_{log}$), parameter accuracy (P: $S_{prm}$), pass@1 (A: $S_{pass}$), and final score (Final: $S_{final}$).}
\begin{tabular}{l|ccccc|ccccc|ccccc|ccccc}
\toprule
& \multicolumn{5}{c|}{Direct, Single-step} & \multicolumn{5}{c|}{Direct, Multi-step} & \multicolumn{5}{c|}{Retrieval, Single-step} & \multicolumn{5}{c}{Retrieval, Multi-step} \\
Model 
& F & L & P & A & Final 
& F & L & P & A & Final 
& F & L & P & A & Final 
& F & L & P & A & Final \\
\midrule
Claude Haiku 4.5 & 0.97 & 0.44 & 0.47 & 0.01 & 0.33 & 0.87 & 0.39 & 0.45 & 0.04 & 0.31 & 1.00 & 0.85 & 0.86 & 0.35 & 0.67 & 0.97 & 0.71 & 0.74 & 0.27 & 0.57 \\
Claude Opus 4.5 & 0.99 & 0.53 & 0.51 & 0.01 & 0.37 & 0.98 & 0.55 & 0.55 & \textbf{0.11} & \textbf{0.42} & 1.00 & 0.87 & 0.90 & 0.41 & \textbf{0.70} & 0.96 & 0.73 & 0.74 & 0.29 & 0.58 \\
Codestral & 0.93 & 0.43 & 0.42 & 0.01 & 0.31 & 0.80 & 0.38 & 0.40 & 0.05 & 0.29 & 0.94 & 0.76 & 0.79 & 0.32 & 0.61 & 0.84 & 0.68 & 0.67 & 0.24 & 0.52 \\
DeepSeek 3.2 & 0.96 & 0.45 & 0.50 & 0.04 & 0.35 & 0.94 & 0.40 & 0.39 & 0.05 & 0.31 & 0.99 & 0.82 & 0.85 & 0.33 & 0.65 & 0.97 & 0.73 & 0.76 & \textbf{0.32} & \textbf{0.59} \\
GLM 4.6 & 0.99 & 0.47 & 0.46 & 0.02 & 0.34 & 0.88 & 0.41 & 0.41 & 0.02 & 0.30 & 1.00 & 0.86 & 0.88 & 0.38 & 0.69 & 0.96 & 0.72 & 0.75 & 0.22 & 0.55 \\
GLM 4 32B & 0.95 & 0.40 & 0.42 & 0.02 & 0.31 & 0.88 & 0.35 & 0.38 & 0.01 & 0.27 & 0.99 & 0.82 & 0.85 & 0.35 & 0.65 & 0.91 & 0.66 & 0.66 & 0.20 & 0.50 \\
GPT 5.2 & 1.00 & 0.48 & 0.54 & \textbf{0.06} & \textbf{0.38} & 0.88 & 0.39 & 0.45 & 0.04 & 0.31 & 1.00 & 0.81 & 0.86 & 0.39 & 0.67 & 0.94 & 0.63 & 0.68 & 0.22 & 0.51 \\
GPT 5.2 Codex & 1.00 & 0.37 & 0.36 & 0.01 & 0.29 & 0.60 & 0.30 & 0.33 & 0.03 & 0.23 & 1.00 & 0.79 & 0.83 & 0.38 & 0.66 & 0.86 & 0.65 & 0.70 & 0.32 & 0.55 \\
Gemini 2.5 Flash & 1.00 & 0.41 & 0.49 & 0.01 & 0.32 & 0.96 & 0.41 & 0.50 & 0.04 & 0.33 & 1.00 & 0.85 & 0.89 & \textbf{0.42} & \textbf{0.70} & 1.00 & 0.75 & 0.77 & 0.27 & 0.58 \\
Gemini 2.5 Pro & 1.00 & 0.50 & 0.54 & 0.05 & \textbf{0.38} & 0.95 & 0.48 & 0.49 & 0.10 & 0.38 & 0.99 & 0.84 & 0.87 & 0.41 & 0.69 & 0.95 & 0.72 & 0.75 & 0.31 & \textbf{0.59} \\
Kimi 2 & 0.99 & 0.47 & 0.48 & 0.01 & 0.34 & 0.94 & 0.50 & 0.46 & 0.03 & 0.35 & 1.00 & 0.83 & 0.87 & 0.37 & 0.67 & 0.96 & 0.70 & 0.73 & 0.24 & 0.55 \\
Llama 3.1 8B & 0.91 & 0.33 & 0.32 & 0.00 & 0.25 & 0.90 & 0.32 & 0.37 & 0.00 & 0.26 & 0.81 & 0.64 & 0.65 & 0.19 & 0.48 & 0.79 & 0.60 & 0.58 & 0.10 & 0.41 \\
Llama 3.3 70B & 0.87 & 0.38 & 0.39 & 0.00 & 0.28 & 0.89 & 0.36 & 0.41 & 0.02 & 0.29 & 0.93 & 0.74 & 0.78 & 0.26 & 0.58 & 0.92 & 0.66 & 0.70 & 0.19 & 0.51 \\
Ministral 14B & 0.39 & 0.15 & 0.17 & 0.00 & 0.12 & 0.46 & 0.18 & 0.21 & 0.01 & 0.15 & 0.81 & 0.67 & 0.68 & 0.25 & 0.52 & 0.69 & 0.48 & 0.49 & 0.16 & 0.37 \\
Qwen3 235B & 1.00 & 0.41 & 0.39 & 0.01 & 0.30 & 0.90 & 0.35 & 0.41 & 0.03 & 0.29 & 0.94 & 0.78 & 0.82 & 0.37 & 0.64 & 0.96 & 0.66 & 0.69 & 0.28 & 0.54 \\
Qwen3 Coder & 1.00 & 0.40 & 0.43 & 0.02 & 0.31 & 0.93 & 0.33 & 0.38 & 0.03 & 0.28 & 0.93 & 0.78 & 0.80 & 0.35 & 0.63 & 0.85 & 0.63 & 0.64 & 0.23 & 0.49 \\
\bottomrule
\end{tabular}
\label{5_1_exp}
\end{table*}

Table~\ref{5_1_exp} presents the evaluation results under direct inference and retrieval-augmented settings across both single-step and multi-step tasks. We observe a consistent and substantial performance improvement from retrieval augmentation across all evaluated models and metrics.

First, retrieval dramatically improves logical correctness ($S_{log}$) and parameter accuracy ($S_{prm}$). For example, Claude Haiku 4.5 improves from 0.44 to 0.85 in $S_{log}$ and from 0.47 to 0.86 in $S_{prm}$ on single-step tasks. Similar trends hold across nearly all models, with gains of +0.30 to +0.40 in both metrics. This suggests that providing in-context demonstrations helps models better identify the correct contract-function pairs and extract structured parameters more accurately.

Second, retrieval leads to a significant increase in end-to-end success rate ($S_{pass}$). Under direct inference, most models achieve near-zero pass@1 (typically below 0.05), indicating that fully correct transaction generation is rare. In contrast, retrieval boosts pass@1 to the 0.30–0.42 range for strong models such as Gemini 2.5 Flash and Claude Opus 4.5. This highlights that retrieval is not merely improving partial correctness, but is critical for achieving executable, fully correct outputs.

Third, the improvements are consistent across both single-step and multi-step tasks, although the latter remains more challenging. While retrieval narrows the gap, multi-step performance still lags behind single-step performance (e.g., 0.70 vs. 0.58 in final score for Claude Opus 4.5), indicating that compositional reasoning over multiple transactions remains a key bottleneck. To further illustrate the specific nature of these reasoning patterns and the qualitative failure modes observed during evaluation, we provide a case study in Appendix \ref{case_of}.

Interestingly, we observe that code-specialized models do not consistently outperform general-purpose models. For instance, GPT 5.2 Codex underperforms its base counterpart in direct inference, suggesting that code-centric pre-training alone is insufficient for this task. Instead, retrieval appears to provide a stronger inductive bias by grounding generation in relevant transaction patterns.

\subsection{Scaling Training Data Yields Consistent Gains with Diminishing Returns (RQ2)}

\begin{figure*}[h!]
\centering
\setlength{\tabcolsep}{2pt}

\begin{tabular}{ccccc}

\includegraphics[width=0.21\textwidth]{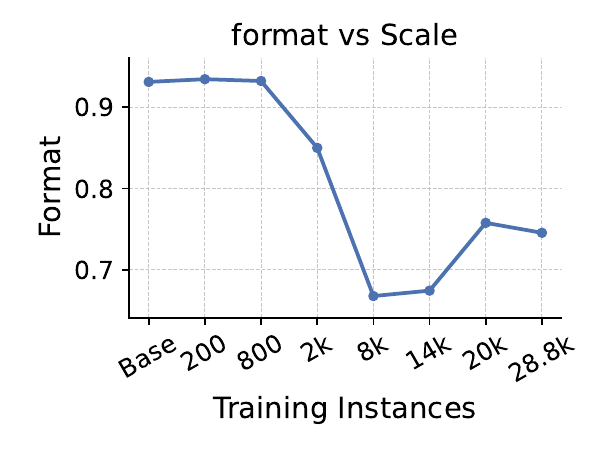} &
\includegraphics[width=0.21\textwidth]{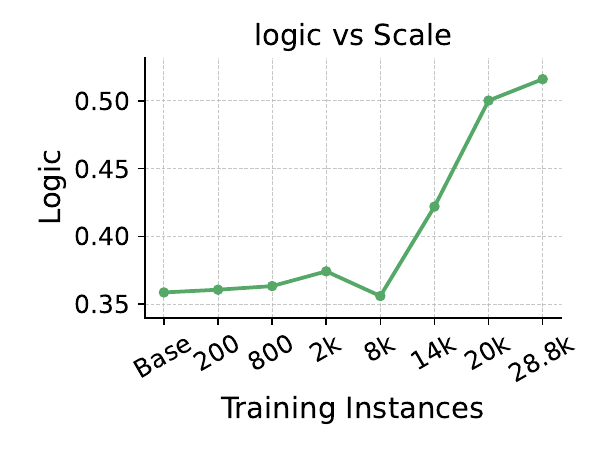} &
\includegraphics[width=0.21\textwidth]{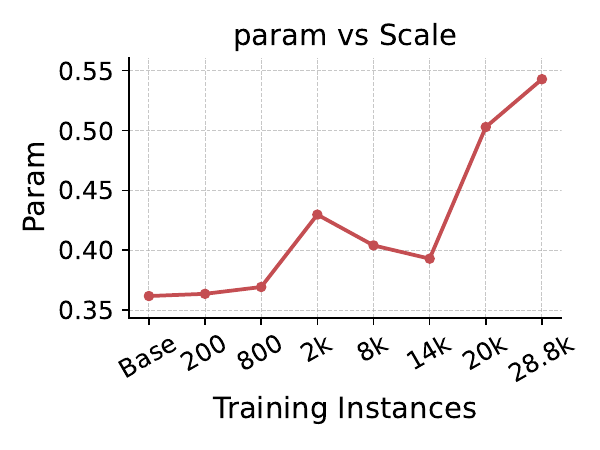} &
\includegraphics[width=0.21\textwidth]{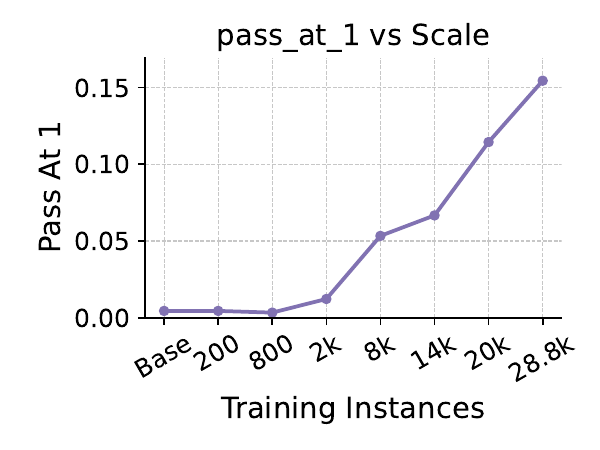} &
\includegraphics[width=0.21\textwidth]{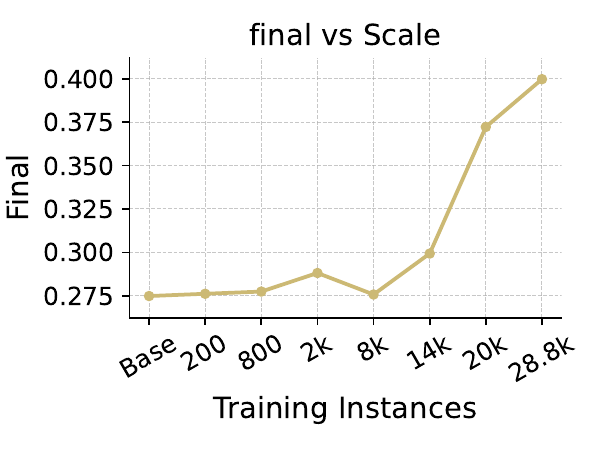} \\

\small
(a) Format &
\small
(b) Logic &
\small
(c) Param &
\small
(d) Pass@1 &
\small
(e) Final \\

\end{tabular}

\caption{
Scaling behavior across dataset sizes.
Each subplot shows performance with increasing training instances.
}

\label{scaling_plots}
\end{figure*}

Figure~\ref{scaling_plots} shows that increasing training data generally improves downstream performance, suggesting that more diverse intent--transaction pairs enhance generalization. The largest gains appear in logic ($S_{log}$) and parameter accuracy ($S_{prm}$), indicating that larger corpora help models better recover contract-function mappings and parameter values. 

We also observe a slight drop in format accuracy ($S_{fmt}$) as data scale increases. This likely reflects a trade-off in dataset complexity: smaller subsets contain simpler and more uniform instances, whereas larger subsets introduce more diverse transaction structures and parameter schemas, making strict JSON adherence harder. Unlike typical saturation behavior, performance continues to improve at larger scales, especially for pass@1 ($S_{pass}$) and the final score. Overall, these results suggest that data quality and diversity are as important as scale for intent-to-transaction generation.

\subsection{Models Exhibit Limited Generalization to Unseen Operation Categories (RQ3)}

\begin{figure}[h!]
    \centering
    \includegraphics[width=400bp]{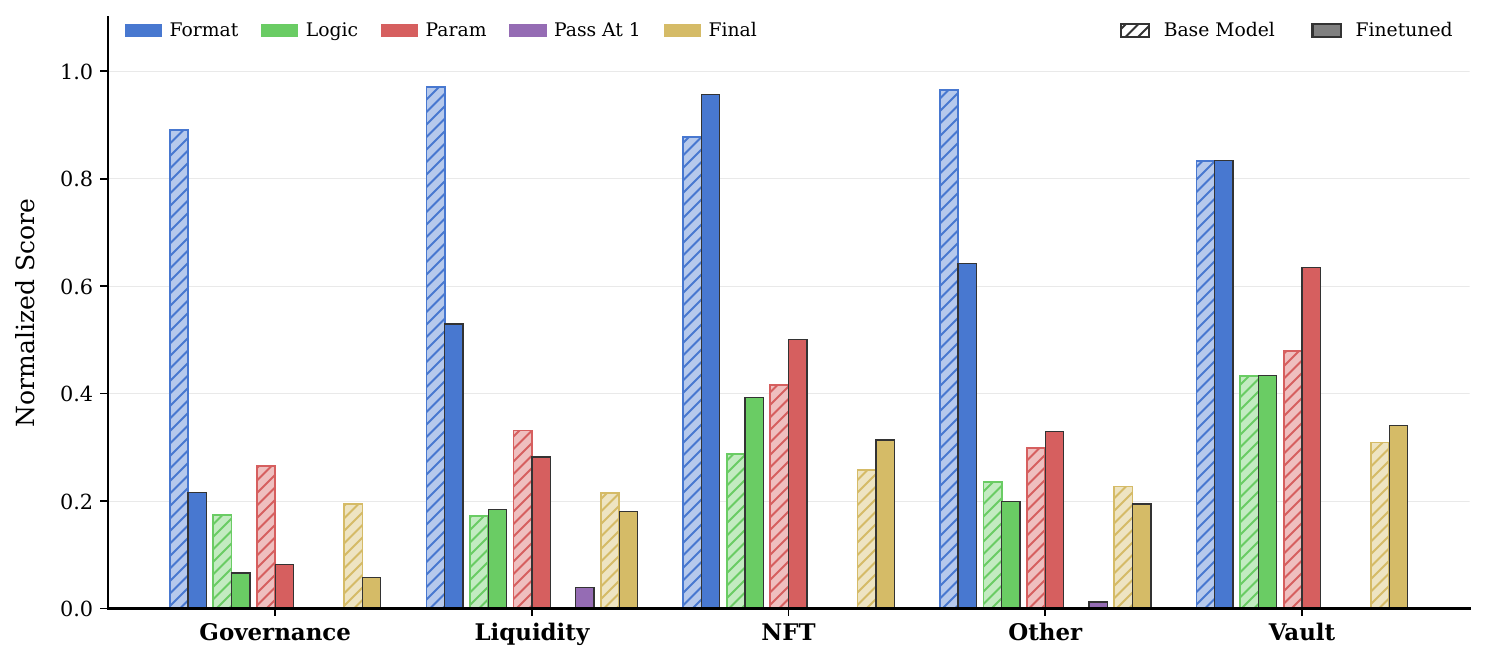}
    \caption{Generalization performance comparison across different categories. The \textit{hatched} bars represent the Base model, while the \textit{solid} colored bars denote cross-category fine-tuned model.}
    \label{genelization_plots}
\end{figure}

Figure~\ref{genelization_plots} compares the base model with the cross-category fine-tuned model on five held-out operation categories. Overall, we observe only limited transfer: fine-tuning does not yield uniform gains on unseen categories, and improvements appear only for a subset of tasks.

In particular, the fine-tuned model performs better on NFT and Vault, where logic, parameter, and final scores all improve to some extent. In contrast, performance on Governance, Liquidity, and Other remains weak or even declines after fine-tuning. One notable effect is that format accuracy can drop substantially on several held-out categories, suggesting that category-specific training may increase output variability without improving general compositional robustness.

These results indicate that cross-category generalization is still a bottleneck. Models trained on common DeFi operations can partially transfer to structurally similar unseen categories, but they do not reliably generalize to rarer or more specialized ones. This gap highlights the need for broader category coverage and more diverse compositional supervision in training data.

\subsection{Execution-based Evaluation Reveals the Potential of Models to Fulfill User Intent (RQ4)}


\begin{table*}[h!]
\centering
\small
\setlength{\tabcolsep}{5pt}
\caption{Simulation execution performance. Exec denotes executability, and StateEq denotes state equivalence with the ground truth. All results are computed over instances with successful ground-truth execution. }
\begin{tabular}{lcccc @{\hspace{1.5em}} lcccc}
\toprule
 & \multicolumn{2}{c}{Direct} & \multicolumn{2}{c}{Retrieval} &  & \multicolumn{2}{c}{Direct} & \multicolumn{2}{c}{Retrieval} \\
\cmidrule(lr){2-3} \cmidrule(lr){4-5} \cmidrule(lr){7-8} \cmidrule(lr){9-10}
Model & Exec & StateEq & Exec & StateEq & Model & Exec & StateEq & Exec & StateEq \\
\midrule
Claude Haiku 4.5 & 0.572 & 0.513 & 0.681 & 0.647  & Gemini 2.5 Flash & 0.662 & \textbf{0.616} & 0.713 & 0.692  \\
Claude Opus 4.5 & 0.513 & 0.472 & 0.729 & \textbf{0.706}  & Gemini 2.5 Pro & \textbf{0.681} & 0.605 & 0.733 & 0.702  \\
Codestral & 0.529 & 0.481 & 0.634 & 0.626  & Kimi 2 & 0.627 & 0.560 & 0.673 & 0.655  \\
DeepSeek 3.2 & 0.542 & 0.480 & 0.669 & 0.635  & Llama 3.1 8B & 0.221 & 0.208 & 0.505 & 0.492  \\
GLM 4.6 & 0.655 & 0.592 & 0.704 & 0.672  & Llama 3.3 70B & 0.323 & 0.310 & 0.592 & 0.573  \\
GLM 4 32B & 0.427 & 0.377 & 0.662 & 0.643  & Ministral 14B & 0.198 & 0.187 & 0.537 & 0.523  \\
GPT 5.2 & 0.618 & 0.541 & \textbf{0.744} & 0.697  & Qwen3 235B & 0.447 & 0.409 & 0.673 & 0.653  \\
GPT 5.2 Codex & 0.437 & 0.394 & 0.720 & 0.697  & Qwen3 Coder & 0.554 & 0.520 & 0.721 & 0.697  \\
\bottomrule
\end{tabular}
\label{5_4_exp}
\end{table*}

Table~\ref{5_4_exp} presents execution-based evaluation results, measuring whether generated transactions are executable (Exec) and whether they achieve state equivalence (StateEq) with the ground truth. This evaluation directly captures whether model outputs can successfully fulfill single-step user intent on-chain, beyond surface-level correctness.

Overall, we observe a substantial gap between syntactic correctness and execution success. Even strong models achieve only moderate performance in the Direct setting, with Exec and StateEq typically ranging between 0.4--0.6. This indicates that generating structurally valid transactions does not guarantee correct on-chain behavior, highlighting the importance of execution-based evaluation.

Retrieval augmentation consistently improves performance across all models. Both Exec and StateEq scores increase significantly, with top models such as GPT~5.2 and Gemini~2.5 Pro achieving over 0.73 in executability and around 0.70 in state equivalence. This suggests that grounding generation in relevant historical transactions or contract interactions helps reduce critical errors and improves end-to-end correctness.

Interestingly, the gap between Exec and StateEq remains non-trivial across settings. Many generated transactions can be executed but still fail to produce the correct state changes, implying that models often approximate the intended behavior without fully matching the ground truth. This further validates the necessity of state-based evaluation rather than relying solely on executability.

Finally, smaller open-source models lag significantly behind frontier models, especially in the Direct setting, but benefit markedly from retrieval. This indicates that while scaling improves execution reliability, external knowledge remains a key factor for robust intent fulfillment.

Overall, these results demonstrate that execution-based evaluation reveals meaningful differences in models' ability to fulfill user intent, and provides a more faithful measure of real-world utility than static matching metrics.
\section{Conclusion}

We presented \textsc{Intent2Tx}, a benchmark for translating natural language intents into structured and executable Ethereum transactions, together with a data construction pipeline, evaluation metrics, and simulation-based execution analysis. Built from real Ethereum mainnet activity, the benchmark captures both common DeFi operations and long-tail categories, enabling systematic study of intent-to-transaction generation in realistic Web3 settings. Our experiments on 16 LLMs show that retrieval augmentation is highly effective, especially for improving logical matching, parameter extraction, and end-to-end transaction quality. We also find that larger fine-tuning datasets improve performance, but the gains depend strongly on data diversity and task complexity rather than scale alone. At the same time, cross-category generalization remains limited: models fine-tuned on common operations transfer only partially to rarer categories such as governance, liquidity, NFT, and vault tasks. Finally, execution-based evaluation demonstrates that exact-match metrics alone substantially underestimate model capability, since many outputs that differ syntactically from the reference can still fulfill the user intent when executed on-chain. We hope \textsc{Intent2Tx} will serve as a useful benchmark for future research on Web3 agents, retrieval-augmented generation, and execution-aware evaluation of structured reasoning systems.




\bibliographystyle{acm}
{
\small
\bibliography{neurips_2026}
}


\appendix

\section{LLMs participating in the evaluation.}
\label{llms_eval}

\begin{table*}[h!]
\centering
\small
\caption{Models evaluated in our benchmark.}
\begin{tabularx}{\textwidth}{l l l >{\raggedright\arraybackslash}X}
\toprule
\textbf{Model} & \textbf{OpenRouter Model Name} & \textbf{Type} & \textbf{Key Strengths / Notes} \\
\midrule

GPT 5.2 & openai/gpt-5.2 & Closed &
Strong reasoning, long context ($\sim$400k), fast inference \\

GPT 5.2 Codex & openai/gpt-5.2-codex & Closed &
Optimized for coding and tool-use workflows \\

Claude Haiku 4.5 & anthropic/claude-haiku-4.5 & Closed &
Fast, cost-efficient lightweight model \\

Claude Opus 4.5 & anthropic/claude-opus-4.5 & Closed &
Top-tier coding and agent performance \\

Gemini 2.5 Flash & google/gemini-2.5-flash & Closed &
Low-latency, cost-efficient multimodal model \\

Gemini 2.5 Pro & google/gemini-2.5-pro & Closed &
Long context ($\sim$1M), strong multimodal reasoning \\

DeepSeek 3.2 & deepseek/deepseek-v3.2 & Open-weight &
Competitive reasoning, low cost, MoE architecture \\

Kimi 2 & moonshotai/kimi-k2-0905 & Closed &
Long-context dialogue, strong bilingual ability \\

Llama 3.1 8B & meta-llama/llama-3.1-8b-instruct & Open &
Small, efficient, widely deployable \\

Llama 3.3 70B & meta-llama/llama-3.3-70b-instruct & Open &
Strong general-purpose open model \\

Qwen3 235B & qwen/qwen3-235b-a22b-2507 & Open-weight &
Large MoE model, strong reasoning performance \\

Qwen3 Coder & qwen/qwen3-coder-next & Open-weight &
Specialized for code generation and reasoning \\

Ministral 14B & mistralai/ministral-14b-2512 & Open &
Efficient small model with good latency-performance \\

Codestral & mistralai/codestral-2508 & Open-weight &
Code-focused model for completion and generation \\

GLM 4.6 & z-ai/glm-4.6 & Closed &
General-purpose Chinese/English model \\

GLM 4 32B & z-ai/glm-4-32b & Open-weight &
Mid-scale deployable model \\

\bottomrule
\end{tabularx}
\label{tab:models}
\end{table*}

\section{Prompts}

\subsection{Guiding model to derive intent from transactions.}
\label{prompt_1}

\begin{footnotesize}
\begin{lstlisting}
You are a Web3 User Intent Generator. Your goal is to reverse-engineer a raw blockchain transaction into a realistic user request.

### Logic: Imagine the User's Persona
1. **For Common Actions (Transfer/Swap/Wrap)**: 
   - Act as a casual user. Use colloquial language.
   - Example: "Send all my ETH to my ledger", "Swap 100 USDT for some Pepe", "Wrap 2 Ether".
   - *Key*: Focus on the "what" and "how much", keep it natural.

2. **For Complex/DeFi Actions (Staking/Liquidity/Vaults)**:
   - Act as a DeFi user. Be more precise about the protocol and the action.
   - You can even provide specific details(contract address, function signature) when the transaction is complex or contract is obscure. However, you should be careful about the extent to which you provide this information.
   - Example: "Deposit 500 USDC into the Aave V3 pool", "Claim my rewards from the Lido dashboard".

3. **For Low-level/Niche/Rare Contract Calls**:
   - Act as a power user or developer. If the contract is rare, include specific IDs or addresses provided in the parameters.
   - If the transaction contains a long hex string (like data or secret) that cannot be simplified, explicitly mention that 'pre-calculated data' or 'specific secret' should be used.
   - Example: "Vote 'Yes' on Proposal #42", "Mint 3 NFTs from the whitelist contract at 0x7f18bb4dd92cf2404c54cba1a9be4a1153bdb078", "Call the emergency withdraw on the vault".

### Core Rules:
- **No Technical Metadata**: Do NOT mention 'bytes', 'calldata' or 'uint256'.
- **Param Integration**: If a parameter looks like a 'recipient', 'amount', or 'token symbol', incorporate it into the sentence.
- **Vary the Tone**: Choose professional, casual, and urgent tones according to the transaction content.
- **Output Only**: Provide the sentence only, no preamble.

### User Persona Examples:
- [Input: Transfer ETH] -> "Give 0.5 ETH to my friend at 0x942a6a136f84d491134b9e03a35dafa701d7ea21"
- [Input: Uniswap Swap] -> "I want to exchange 1000 USDC for as much LINK as possible."
- [Input: Niche Mint] -> "Mint 5 tokens from the contract 0x7f18bb4dd92cf2404c54cba1a9be4a1153bdb078 on the mainnet."
\end{lstlisting}
\end{footnotesize}

\subsection{System prompt for intent to transaction task.}
\label{prompt_2}

\begin{footnotesize}
\begin{lstlisting}
You are a DeFi execution planner.
Your task is to convert a user's high-level intent into a SINGLE on-chain action represented as a JSON object.
Output requirements:
- Output JSON only, no explanations.
- Follow the exact schema defined below.
- For each parameter, you must provide both its Solidity type and the specific value inferred from the intent.
JSON schema:\n"
{\n"
  "contract": string,              // contract name (e.g., "UniswapV2Router02")
  "contract_address": string,      // contract address (e.g., "0x7a250d5630b4cf539739df2c5dacabf6959a1e")
  "function": string,              // function name (e.g., "transfer")   
  "params": {                      // map of parameters
    "<param_name>": {\n'
      "type": string,              // Solidity type (e.g., "address", "uint256")
      "val": any                   // specific value inferred from the intent
    }\n'
    ... // more parameters if needed 
  },
  "value": float                   // Ether amount sent with the transaction (in ETH)
}
\end{lstlisting}
\end{footnotesize}

\subsection{Address book provided to LLMs}
\label{address_book}

We provide an address book to all models at inference time to ensure a fair and controlled evaluation. The address book contains mappings between protocol names and on-chain addresses, which are necessary for constructing executable transactions but are not the primary focus of our benchmark.

Without this information, models—especially smaller ones—would be penalized for memorization or hallucination errors unrelated to the core task of intent understanding and transaction composition. By standardizing access to such external knowledge, we isolate the models' reasoning and planning capabilities.

This design follows a common practice in tool-use and agent evaluation, where auxiliary resources are provided to reduce irrelevant variance and improve comparability across models of different scales.

\begin{footnotesize}
\begin{lstlisting}

You may refer to the following address book for commonly used Ethereum protocol addresses. This can help you avoid mistakes when inputting contract addresses for popular tokens and protocols.

## Additional Information: Helpful Ethereum Address Book

This address book provides a mapping between common protocol names (just a description, not contract name), token symbols, and their on-chain addresses.

### Tokens
| Name / Symbol | Contract Address |
| :--- | :--- |
| **USDT** | `0xdAC17F958D2ee523a2206206994597C13D831ec7` |
| **USDC** | `0xA0b86991c6218b36c1d19D4a2e9Eb0cE3606eb48` |
| **DAI** | `0x6B175474E89094C44Da98b954EedeAC495271d0F` |
| **WBTC** | `0x2260FAC5E5542a773Aa44fBCfeDf7C193bc2C599` |
| **WETH** | `0xC02aaA39b223FE8D0A0e5C4F27eAD9083C756Cc2` |
| **PYUSD** | `0x6c3ea9036406852006290770BEdFcAbA0e23A0e8` |
| **LINK** | `0x514910771af9ca656af840dff83e8264ecf986ca` |
| **UNI** | `0x1f9840a85d5af5bf1d1762f925bdaddc4201f984` |
| **fwWETH** | `0xa250cc729bb3323e7933022a67b52200fe354767` |
| **SPX** | `0xe0f63a424a4439cbe457d80e4f4b51ad25b2c56c` |
| **PAXG** | `0x45804880de22913dafe09f4980848ece6ecbaf78` |
| **AAVE** | `0x7fc66500c84a76ad7e9c93437bfc5ac33e2ddae9` |
| **ENA** | `0x57e114b691db790c35207b2e685d4a43181e6061` |
| **XAUt** | `0x68749665ff8d2d112fa859aa293f07a622782f38` |
| **CRV** | `0xd533a949740bb3306d119cc777fa900ba034cd52` |
| **USDS** | `0xdc035d45d973e3ec169d2276ddab16f1e407384f` |

### DEX / Routers
| Name / Symbol | Contract Address |
| :--- | :--- |
| **Uniswap_V2_Router** | `0x7a250d5630B4cF539739dF2C5dAcb4c659F2488D` |
| **Uniswap_V3_Router_2** | `0x68b3465833fb72A70ecDF485E0e4C7bD8665Fc45` |
| **Universal_Router** | `0x3fc91a3afd70395cd496c647d5a6cc9d4b2b7fad` |
| **SushiSwap_Router** | `0xd9e1cE17f2641f24aE83637ab66a2cca9C378B9F` |
| **1inch_Aggregation_Router_V5** | `0x1111111254EEB25477B68fb85Ed929f73A960582` |
| **1inch_Aggregation_Router_V6** | `0x111111125421cA6dc452d289314280a0f8842A65` |
| **0x_Protocol_V4** | `0xDef1C0ded9bec7F1a1670819833240f027b25EfF` |
| **Curve_Router** | `0x99a58482BD75cbab83b27EC03CA68fF489b5788f` |
| **Balancer_V2_Vault** | `0xBA12222222228d8Ba445958a75a0704d566BF2C8` |
| **Kyberswap_V2** | `0x6131B5fae19EA4f9D964eAc0408E4408b66337b5` |

### Lending / Staking
| Name / Symbol | Contract Address |
| :--- | :--- |
| **Aave_V2_Lending_Pool** | `0x7d2768dE32b0b80b7a3454c06BdAc94A69DDc7A9` |
| **Aave_V3_Lending_Pool** | `0x87870Bca3F3fD6335C3F4ce8392D69350B4fA4E2` |
| **Lido_stETH** | `0xae7ab96520DE3A18E5e111B5EaAb095312D7fE84` |
| **Rocket_Pool_RETH** | `0xae78736Cd615f374D3085123A210448E74Fc6393` |
| **Spark_V1_Lending_Pool** | `0xC13e21B648A5Ee794902342038FF3aDAB66BE987` |
| **Morpho_Blue_V2** | `0xBBBBBbbBBb9cC5e90e3b3Af64bdAF62C37EEFFCb` |
| **Compound_V2** | `0x3d9819210A31b4961b30EF54bE2aeD79B9c9Cd3B` |
| **Compound_V3_USDC_Market** | `0xc3d688B66703497DAA19211EEdff47f25384cdc3` |
| **Compound_V3_WETH_Market** | `0xA17581A9E3356d9A858b789D68B4d866e593aE94` |

### NFT
| Name / Symbol | Contract Address |
| :--- | :--- |
| **Opensea_Seaport** | `0x00000000000000ADc04C56Bf30aC9d3c0aAF14dC` |
| **Blur_Blend** | `0x29469395eAf6f95920E59F858042f0e28D98a20B` |
| **CryptoPunks** | `0xb47e3cd837dDF8e4c57F05d70Ab865de6e193BBB` |

### Bridges / Infrastructure
| Name / Symbol | Contract Address |
| :--- | :--- |
| **Across_Spoke_Pool** | `0x5c7BC2d53d245f5021E17E266bb792010103510c` |
| **Optimism_L1_Standard_Bridge** | `0x99C9fc46f92E8a1c0deC1b1747d010903E884bE1` |
| **Arbitrum_L1_Standard_Bridge** | `0x8315177aB297bA92A06054cE80a67Ed4DBd7ed3a` |
| **Base_L1_Standard_Bridge** | `0x3154Cf16ccdb4C6d922629664174b904d80F2C35` |
| **Polygon_RootChainManager** | `0xA0c68C638235ee32657e8f720a23ceC1bFc77C77` |
| **Stargate_Route** | `0x8731d54E9D02c286767d56ac03e8037C07e01e98` |
| **Hop_Ethereum_Bridge** | `0xb8901acB165ed027E32754E0FFe830802919727f` |
| **Synapse_Bridge** | `0x2796317b0fF8538F253012862c06787Adfb8cEb6` |
| **Celer_cBridge_V2** | `0x5427FEFA711Eff984124bFBB1AB6fbf5E3DA1820` |
| **zkSync_Era_L1_DiamondProxy** | `0x32400084C286CF3E17e7B677ea9583e60a000324` |

\end{lstlisting}
\end{footnotesize}

\section{Data Example}
\label{data_example}

To provide a concrete understanding of the \textsc{Intent2Tx} benchmark, we present representative examples illustrating the mapping from high-level natural language intents to structured, executable on-chain calls. Each instance in our dataset contains metadata (omitted here), a user intent and a corresponding JSON-formatted ground truth, which includes the contract name, address, function signature, and typed parameters.

\begin{footnotesize}
\begin{lstlisting}
------------------------------- Example 1 -----------------------------------

Input: User intent: Transfer 4.399312393802940721 LPT to 0x0D0707963952f2FBA59dD06f2B425ace40B492FE.

Output: 
{
  "contract": "LivepeerToken",
  "contract_address": "0x58b6a8a3302369daec383334672404ee733ab239",
  "function": "transfer",
  "params": {
    "_to": {
      "type": "address",
      "val": "0x0d0707963952f2fba59dd06f2b425ace40b492fe"
    },
    "_value": {
      "type": "uint256",
      "val": "4399312393802940721"
    }
  },
  "value": 0
}

------------------------------- Example 2 -----------------------------------

Input: User intent: Burn my NFT position with ID 950002.

Output:
{
  "contract": "NonfungiblePositionManager",
  "contract_address": "0xc36442b4a4522e871399cd717abdd847ab11fe88",
  "function": "burn",
  "params": {
    "tokenId": {
      "type": "uint256",
      "val": "950002"
    }
  },
  "value": 0
}

------------------------------- Example 3 -----------------------------------

Input: User intent: Mint an NFT to my wallet 0xa17a980c0f540489e75664b32cb0f9cb17a4940f from the contract at 0xa9ab98cb65a9c522a196823ffd04dd3f7276c19f, using metadata from ipfs://QmSkQEgqsPSRN5b6nXf7TW6NyJFpEgDJXV9NF6QjvwNAun/0.after that Create a new ERC1155 token with the URI "ipfs://QmSfpSWBYbTiRpdXeG1mdkTudg8UYzc1oC6Vr85CLCGdP3/0" and mint 1 of these tokens to 0xa17a980c0f540489e75664b32cb0f9cb17a4940f.

Output: 
[
  {
    "contract": "ERC721TL",
    "contract_address": "0xa9ab98cb65a9c522a196823ffd04dd3f7276c19f",
    "function": "mint",
    "params": {
      "recipient": {
        "type": "address",
        "val": "0xa17a980c0f540489e75664b32cb0f9cb17a4940f"
      },
      "uri": {
        "type": "string",
        "val": "ipfs://QmSkQEgqsPSRN5b6nXf7TW6NyJFpEgDJXV9NF6QjvwNAun/0"
      }
    },
    "value": 0
  },
  {
    "contract": "ERC1155TL",
    "contract_address": "0x1b10f094efb8bc1defb12af23313d2e45054a25d",
    "function": "createToken",
    "params": {
      "newUri": {
        "type": "string",
        "val": "ipfs://QmSfpSWBYbTiRpdXeG1mdkTudg8UYzc1oC6Vr85CLCGdP3/0"
      },
      "addresses": {
        "type": "address[]",
        "val": [
          "0xa17a980c0f540489e75664b32cb0f9cb17a4940f"
        ]
      },
      "amounts": {
        "type": "uint256[]",
        "val": [
          "1"
        ]
      }
    },
    "value": 0
  }
]

\end{lstlisting}
\end{footnotesize}

The examples above showcase several key challenges addressed by \textsc{Intent2Tx}:

Precision and Unit Normalization: In \texttt{Example 1}, the model must accurately convert a floating-point token amount (4.399... LPT) into its base unit (\textit{wei}) equivalent as a \texttt{uint256} string, requiring knowledge of the token's decimal precision (18 decimals for LPT).

Protocol-Specific Mapping: \texttt{Example 2} demonstrates the transition from domain-specific jargon (``burn my NFT position'') to the specific logic of the \textit{NonfungiblePositionManager} contract, identifying the correct function (\texttt{burn}) and the associated \texttt{tokenId}.

Multi-step Logic and Composition: \texttt{Example 3} illustrates a complex, multi-goal intent. The model must decompose the request into a sequence of atomic operations—first minting an ERC721 token and then creating a new ERC1155 token. This requires not only correct parameter extraction (e.g., handling \texttt{ipfs} URIs) but also strategic planning to maintain the execution order.

Complex Type Handling: As seen in the output for the ERC1155 creation, the benchmark requires models to format parameters into complex Solidity types, such as \texttt{address[]} and \texttt{uint256[]} arrays, ensuring they are ready for ABI encoding.

\section{Case of LLMs generating transactions}
\label{case_of}

To provide a deeper understanding of the model's reasoning capabilities and failure modes, we present several qualitative examples across different model families and task complexities. These cases are extracted from our experimental evaluations, showcasing both successful logical mappings and representative errors in numerical precision, protocol selection, and address handling.

\begin{footnotesize}
\begin{lstlisting}
---------------------- Claude Haiku 4.5 : Single-step -----------------------

Input: User intent: Swap 0.0534 ETH for at least 153.71M units of token from address 0x5148e5CfDfDF76987cD1c6033bbCD52DF0A8d75F, sending them to 0xb86c24408f7ce28fc4742493aeb9776f163a19bc.

Output:
{
  "contract": "UniswapV2Router02",
  "contract_address": "0x7a250d5630b4cf539739df2c5dacb4c659f2488d",
  "function": "swapExactETHForTokens",
  "params": {
    "amountOutMin": {
      "type": "uint256",
      "val": "1537119548300416251509265942"
    },
    "path": {
      "type": "address[]",
      "val": [
        "0xc02aaa39b223fe8d0a0e5c4f27ead9083c756cc2",
        "0x5148e5cfdfdf76987cd1c6033bbcd52df0a8d75f"
      ]
    },
    "to": {
      "type": "address",
      "val": "0xb86c24408f7ce28fc4742493aeb9776f163a19bc"
    },
    "deadline": {
      "type": "uint256",
      "val": "1764335997"
    }
  },
  "value": 0.0534
}

LLM_Output:
{
  "contract": "UniswapV2Router02",
  "contract_address": "0x7a250d5630b4cf539739df2c5dacb4c659f2488d",
  "function": "swapExactETHForTokens",
  "params": {
    "amountOutMin": {
      "type": "uint256",
      "val": "153710000000000000000000000"
    },
    "path": {
      "type": "address[]",
      "val": [
        "0xc02aaa39b223fe8d0a0e5c4f27ead9083c756cc2",
        "0x5148e5CfDfDF76987cD1c6033bbCD52DF0A8d75F"
      ]
    },
    "to": {
      "type": "address",
      "val": "0xb86c24408f7ce28fc4742493aeb9776f163a19bc"
    },
    "deadline": {
      "type": "uint256",
      "val": "1742220567"
    }
  },
  "value": 0.0534
}

------------------------ Llama 3.1 8B : Single-step -------------------------

Input: User intent: Swap 0.015874411512787778 units of the token at 0x13919f59286578600321f83941369e319dbc0bc5 for at least 1.654878062516271904 ETH, and send it to 0x1af8008c5f5bca43305a3d4104d5e1802c49814f. This needs to be done before May 5, 2025 at 1:26:51 PM GMT.

Output:
{
  "contract": "UniswapV2Router02",
  "contract_address": "0x7a250d5630b4cf539739df2c5dacb4c659f2488d",
  "function": "swapExactTokensForETH",
  "params": {
    "amountIn": {
      "type": "uint256",
      "val": "15874411512787778"
    },
    "amountOutMin": {
      "type": "uint256",
      "val": "1654878062516271904"
    },
    "path": {
      "type": "address[]",
      "val": [
        "0x13919f59286578600321f83941369e319dbc0bc5",
        "0xc02aaa39b223fe8d0a0e5c4f27ead9083c756cc2"
      ]
    },
    "to": {
      "type": "address",
      "val": "0x1af8008c5f5bca43305a3d4104d5e1802c49814f"
    },
    "deadline": {
      "type": "uint256",
      "val": "1746274011"
    }
  },
  "value": 0
}

LLM_Output:
{
  "contract": "UniswapV2Router02",
  "contract_address": "0x7a250d5630B4cF539739dF2C5dAcb4c659F2488D",
  "function": "swapExactETHForTokens",
  "params": {
    "amountIn": {
      "type": "uint256",
      "val": "0.015874411512787778"
    },
    "minAmountOut": {
      "type": "uint256",
      "val": "1.654878062516271904"
    },
    "path": {
      "type": "address[]",
      "val": [
        "0x13919f59286578600321f83941369e319dbc0bc5"
      ]
    },
    "to": {
      "type": "address",
      "val": "0x1af8008c5f5bca43305a3d4104d5e1802c49814f"
    },
    "deadline": {
      "type": "uint256",
      "val": "1649146801"
    }
  },
  "value": "0.0"
}

----------------------- Claude Opus 4.5 : Multi-step ------------------------

Input: User intent: Send 167.659520753773760851 Bounce to 0xf5a7b9617422fd252cf1cd9fbc69f42645ad0ead. and Send 15,107,747.49 SPELL to 0xf5a7b9617422fd252cf1cd9fbc69f42645ad0ead.

Output:
[{"contract": "BounceAuctionToken", "contract_address": "0xa9b1eb5908cfc3cdf91f9b8b3a74108598009096", "function": "transfer", "params": {"recipient": {"type": "address", "val": "0xf5a7b9617422fd252cf1cd9fbc69f42645ad0ead"}, "amount": {"type": "uint256", "val": "167659520753773760851"}}, "value": 0.0}, {"contract": "Spell", "contract_address": "0x090185f2135308bad17527004364ebcc2d37e5f6", "function": "transfer", "params": {"to": {"type": "address", "val": "0xf5a7b9617422fd252cf1cd9fbc69f42645ad0ead"}, "amount": {"type": "uint256", "val": "15107747493136224938278790"}}, "value": 0.0}]

LLM_Output:
[{"contract": "BounceAuctionToken", "contract_address": "0xa9b1eb5908cfc3cdf91f9b8b3a74108598009096", "function": "transfer", "params": {"recipient": {"type": "address", "val": "0xf5a7b9617422fd252cf1cd9fbc69f42645ad0ead"}, "amount": {"type": "uint256", "val": "167659520753773760851"}}, "value": 0.0}, {"contract": "SpellToken", "contract_address": "0x090185f2135308bad17527004364ebcc2d37e5f6", "function": "transfer", "params": {"to": {"type": "address", "val": "0xf5a7b9617422fd252cf1cd9fbc69f42645ad0ead"}, "amount": {"type": "uint256", "val": "15107747000000000000000000"}}, "value": 0.0}]

----------------------- Ministral 14B : Multi-step --------------------------

Input: User intent: I want to resolve the dispute game.and Resolve up to 512 claims, starting from claim index 0, on the Permissioned Dispute Game contract.

Output: 
[{"contract": "PermissionedDisputeGame", "contract_address": "0xb86bf0d32b4676bca2053cca5fc6a512bded4b9a", "function": "resolve", "params": {}, "value": 0.0}, {"contract": "PermissionedDisputeGame", "contract_address": "0xee17cdba5a1490c4f5fe76d906b838832bc0b099", "function": "resolveClaim", "params": {"_claimIndex": {"type": "uint256", "val": "0"}, "_numToResolve": {"type": "uint256", "val": "512"}}, "value": 0.0}]

LLM_Output: 
[ { "contract": "PermissionedDisputeGame", "contract_address": "0x0000000000000000000000000000000000000000", // Placeholder; replace with actual contract address if known "function": "resolve", "params": {}, "value": 0.0 }, { "contract": "PermissionedDisputeGame", "contract_address": "0x0000000000000000000000000000000000000000", // Placeholder; replace with actual contract address if known "function": "resolveClaim", "params": { "_claimIndex": { "type": "uint256", "val": "0" }, "_numToResolve": { "type": "uint256", "val": "512" } }, "value": 0.0 } ]

\end{lstlisting}
\end{footnotesize}

By comparing the ground truth with the outputs from various models, we identify four primary failure modes:

Function Selection and Parameter Formatting: The \textit{Llama 3.1 8B} case illustrates a logic reversal error. The model mistakenly selected \texttt{swapExactETHForTokens} for an intent that required selling a specific token for ETH (\texttt{swapExactTokensForETH}). Furthermore, it failed to perform the floating-to-integer conversion, passing a decimal string into a \texttt{uint256} field, which is syntactically invalid for ABI encoding.

Address Hallucination and Placeholder Usage: \textit{Ministral 14B} exhibits a "placeholder hallucination" pattern in multi-step tasks. While it correctly understands the sequence of operations (resolving a dispute and its claims), it fails to retrieve or infer the specific contract addresses from its context, defaulting to the null address (\texttt{0x00...00}). This highlights the model's dependency on high-quality Retrieval-Augmented Generation (RAG) for long-tail protocol interaction.

Multi-step Decomposition Hurdles: In the \textit{Claude Opus} multi-step example, the model successfully decomposes the intent into two separate \texttt{transfer} calls. However, the accumulation of minor numerical discrepancies across steps remains a challenge for even state-of-the-art models, suggesting that current LLMs lack the internal deterministic execution environment needed for high-stakes financial operations.


\end{document}